\pdfoutput=1

\documentclass[11pt]{article}

\usepackage[final]{acl}

\usepackage{times}
\usepackage{latexsym}
\usepackage{verbatim}
\usepackage{amsmath}
\usepackage{cleveref}
\usepackage{seqsplit}

\definecolor{cvprblue}{rgb}{0.21,0.49,0.74}
\hypersetup{
    colorlinks=true,
    citecolor=cvprblue
}

\usepackage[T1]{fontenc}

\usepackage[utf8]{inputenc}

\usepackage{microtype}

\usepackage{inconsolata}

\usepackage{graphicx}
\usepackage{booktabs}
\usepackage[table]{xcolor}
\usepackage{makecell}
\usepackage{multirow}
\usepackage{amsmath} 
\usepackage{tabularx}
\usepackage{subfigure}
\usepackage{marvosym}

\title{LM-Searcher: Cross-domain Neural Architecture Search with LLMs via Unified Numerical Encoding}

\author{Yuxuan Hu\textsuperscript{1},\quad Jihao Liu\textsuperscript{1},\quad Ke Wang\textsuperscript{1},\quad Jinliang Zheng\textsuperscript{3,4},\quad Weikang Shi\textsuperscript{1},\quad \\ \textbf{Manyuan Zhang\textsuperscript{1},\quad Qi Dou\textsuperscript{2},\quad Rui Liu\textsuperscript{1},\quad Aojun Zhou\textsuperscript{1}$^{\textrm{\Letter}}$,\quad Hongsheng Li\textsuperscript{1,4,5}$^{\textrm{\Letter}}$}\\
$^1$CUHK MMLab\quad $^2$CUHK CURI\quad $^3$Tsinghua University\\$^4$Shanghai AI Laboratory\quad  $^5$CPII under InnoHK\\
$^{\textrm{\Letter}}$Corresponding author\\
\texttt{\{huyuxuan621,aojunzhou\}@gmail.com}\quad \texttt{hsli@ee.cuhk.edu.hk}}

\begin{document}
\maketitle
\begin{abstract}
Recent progress in Large Language Models (LLMs) has opened new avenues for solving complex optimization problems, including Neural Architecture Search (NAS). However, existing LLM-driven NAS approaches rely heavily on prompt engineering and domain-specific tuning, limiting their practicality and scalability across diverse tasks. In this work, we propose LM-Searcher, a novel framework that leverages LLMs for cross-domain neural architecture optimization without the need for extensive domain-specific adaptation. Central to our approach is NCode, a universal numerical string representation for neural architectures, which enables cross-domain architecture encoding and search. We also reformulate the NAS problem as a ranking task, training LLMs to select high-performing architectures from candidate pools using instruction-tuning samples derived from a novel pruning-based subspace sampling strategy. Our curated dataset, encompassing a wide range of architecture-performance pairs, encourages robust and transferable learning. Comprehensive experiments demonstrate that LM-Searcher achieves competitive performance in both in-domain (e.g., CNNs for image classification) and out-of-domain (e.g., LoRA configurations for segmentation and generation) tasks, establishing a new paradigm for flexible and generalizable LLM-based architecture search. The datasets and models will be released at \url{https://github.com/Ashone3/LM-Searcher}.
\end{abstract}

\section{Introduction}
\label{sec:intro}
Recent advances in Large Language Models (LLMs) have demonstrated remarkable potential in solving complex optimization problems, including competitive programming~\cite{openai2024o1}, the Traveling Salesman Problem (TSP)~\cite{yang2023large}, and even Neural Architecture Search (NAS)~\cite{zheng2023can,nasir2024llmatic}. Among these applications, NAS focuses on discovering optimal neural architectures that maximize performance in a given domain. Previous approaches, such as GENIUS and LLMatic, leverage GPT-4~\cite{openai2024gpt4} to generate task-specific architectures, either in code or natural language form.

These preliminary attempts focus on designing prompts to instruct off-the-shelf LLMs for the architectural design of image classification models. However, prompt engineering-based approaches require extensive domain-specific expertise and manual prompt tuning to handle different search spaces and tasks, making them less practical in real-world scenarios. For instance, while GPT-4o can effectively identify better architectures within a simple classification search space such as CIFAR-10, it fails to surpass the random search approach when applied to larger search spaces like Diffusion Transformer~\cite{peebles2023scalable} for image generation.

In this paper, we ask the following question: \textit{Can LLMs be trained as general-purpose search models capable of optimizing neural architectures across diverse domains, without the need for domain-specific tuning?}

\begin{figure}[t]
  \centering
  \includegraphics[width=1.0\linewidth]{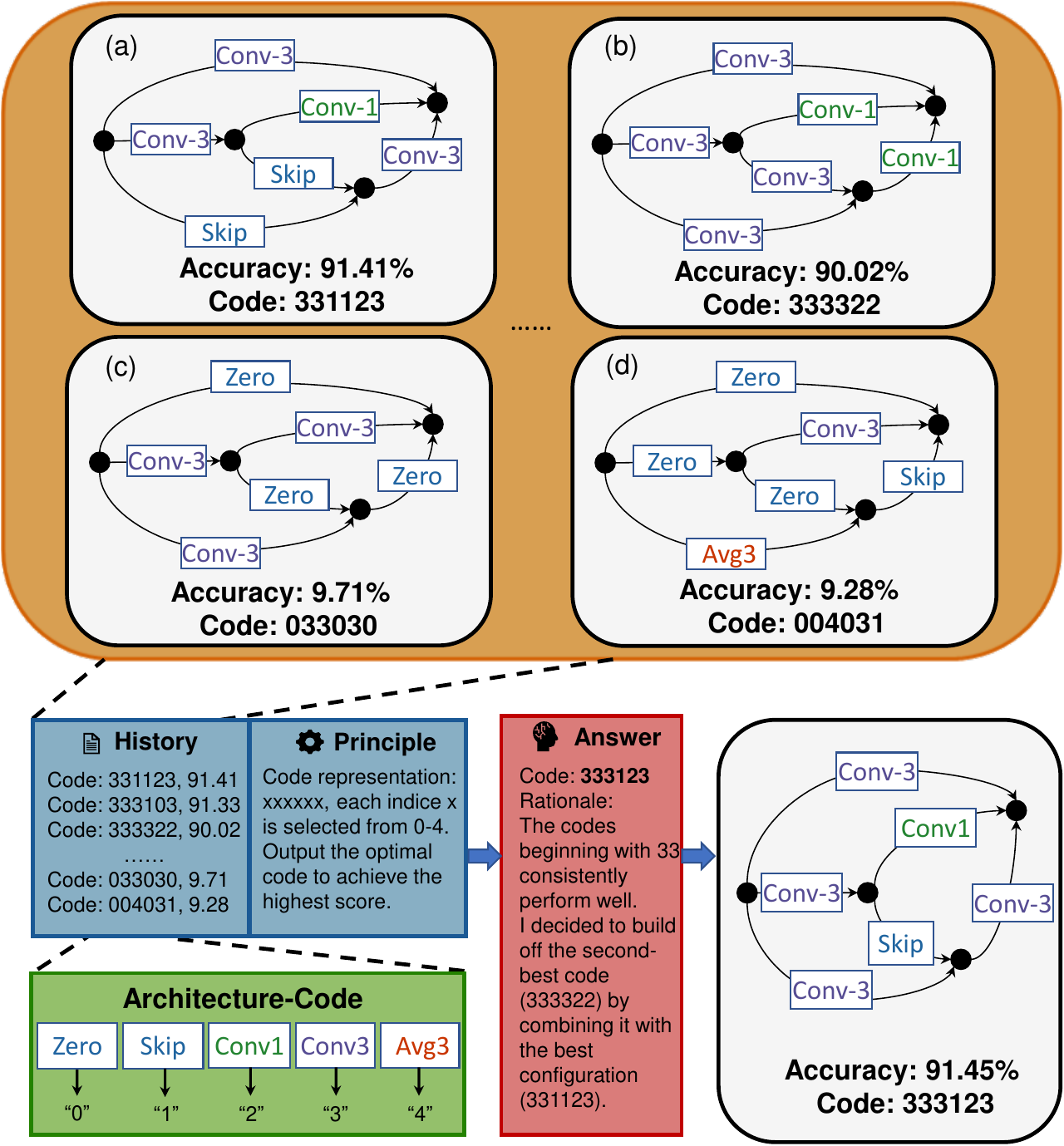}
  \caption{Architecture configurations are encoded as simplified numerical codes and provided to the LLM, which predicts the most promising candidate. This code is then decoded back into a neural architecture for performance evaluation.}
  \label{fig:intro_fig1}
\end{figure}

To answer this question, we introduce LM-Searcher, a novel framework that leverages LLM's reasoning and optimization capabilities for exploring neural architecture, which transforms neural architectures into task-agnostic universal representations, enabling cross-domain search using either off-the-shelf LLMs or a fine-tuned, search-specific LLM. Specifically, architectures are first represented as numerical strings termed \textbf{NCode} for simplicity, and inputed into the LLM for architecture encoding. As illustrated in Fig~\ref{fig:intro_fig1}, architecture configurations are represented by their sub-module indices in the search space. This encoding method unifies tasks across different domains by representing them as combinatorial optimization problems. For example, in NAS-Bench-201~\cite{dong2020bench}, each architecture is built from "cell" units with five configurable operations, encoded as: ["0" (zeroize), "1" (skip connection), "2" (1×1 conv)...]. Unique architectures are represented by combinations of these operation codes.

Such numerical encoding of architectures not only bridges the gap between different architecture domains, but also directs the model's attention towards analyzing inter-architecture relations based on their performance, minimizing distractions from domain-specific characteristics. Subsequently, we extract architecture-performance metadata from established NAS benchmarks~\cite{ying2019bench,zela2020surrogate},
which contains up to \(10^{18}\) possible \textit{NCode-accuracy} pairs (e.g., [NCode: 333123, Accuracy: 91.45\%]). A straightforward approach is to directly use the accuracy of known historical \textit{NCode-accuracy} pairs to predict better-performing architectures.

However, due to the vastness of the neural architecture search space and the large-scale possible combinations of historical and predicted architectures, it is challenging for LLMs to generate the optimal architecture directly. To mitigate this challenge, we reformulate the task as a ranking problem rather than a generation problem, i.e., the LLM is trained to always choose the next best architecture from a candidate pool. 

To effectively construct instruction-tuning samples that conform to our proposed encoding scheme and enhance the LLM's capabilities to explore neural architectures, we introduce a novel pruning-based data curation strategy. This strategy prunes a fully-connected supernet to obtain various sub-networks, following the approach of~\cite{pham2018efficient}. Prior works on pruning~\cite{liu2018rethinking} and weight-sharing NAS~\cite{xu2019pc} have demonstrated that architectures within a reduced sub-network can converge faster and achieve performance comparable to, or even better than, their larger counterparts. Based on this, we sample 100–200 architectures from the same subspace for both the history (NCode-performance) and candidate (NCode) sets, and select the highest-performing architecture as the ground truth. In total, we obtain a 228k instruction-tuning dataset for LLM training.

Training on this carefully curated dataset allows our LLM to learn robust and transferable architecture-performance patterns. As a result, our LM-Searcher can effectively search for architectures not only in image classification tasks but also in previously unseen search spaces. Through comprehensive evaluation of LM-Searcher across diverse tasks without specific tuning, we demonstrate the effectiveness of the proposed LM-Searcher model in in-domain CNN architecture search for image classification (e.g. MobileNetV2) and out-of-domain model design (e.g., LoRA configuration for visual segmentation and generation tasks), respectively. 

Our main contributions can be summarized as follows:

\begin{itemize}
    \item To the best of our knowledge, we introduce \textbf{NCode}, the first cross-domain numerical string representation for neural architectures, enabling architecture optimization across diverse domains.
    
    \item We develop LM-Searcher, a task-agnostic architecture search model trained using a novel pruning-based subspace sampling data curation mechanism.
    
    \item We demonstrate the effectiveness of LM-Searcher across both in-domain CNN architecture search and out-of-domain model design tasks, highlighting its flexibility and generalizability.
\end{itemize}

\section{Related Work}

\subsection{LLMs Solving Complex Problems}
Large language models (LLMs), pre-trained on massive datasets, have demonstrated remarkable proficiency in few-shot learning scenarios~\citep{brown2020language}. The field has witnessed rapid advancements in LLM development, with notable contributions including models like LLaMA~\citep{touvron2023llama}, PaLM~\citep{chowdhery2023palm}, and GPT-4~\citep{openai2024gpt4}. Subsequent research has focused on fine-tuning these models, particularly open-source models like LLaMA, on specialized or synthesized datasets. This has led to significant performance improvements in areas such as code generation~\citep{roziere2023code} and mathematical reasoning~\citep{wang2024mathcoder,goutora}. Our work builds upon this trend by exploring the potential of LLMs for the complex task of neural architecture search (NAS). Our approach leverages a carefully curated dataset derived from existing NAS benchmark. This allows LM-Searcher to achieve strong cross domain generalization in NAS tasks, setting a new benchmark for LLM performance in this area.

\subsection{Architecture Search with LLMs}
LLMs have shown promise in supporting model optimization tasks. \cite{jawahar2023llm} utilized GPT-4~\citep{openai2024gpt4} to estimate the performance of different architectures, using these estimations to guide the initialization of NAS. Similarly, EvoPrompting~\citep{chen2024evoprompting} and LLMatic~\cite{nasir2024llmatic} leverage LLM to perform NAS at the code level. Other studies, such as those by~\cite{song2024position} and~\cite{liu2024large}, have explored the potential of LLMs for black-box optimization in the context of NAS. GENIUS~\citep{zheng2023can} employs GPT-4 to suggest model architectures during the search phase. However, these methods often suffer from limitations such as the need for extensive manual prompt engineering, making them difficult to transfer to different tasks. In contrast, our approach treats the LLM as a general-purpose search model for NAS by introducing a universal representation for encoding architectures across diverse tasks and a pruning-based data construction technique for training the LLM, enabling cross-domain generalization without the need for specific tuning.

\begin{figure*}[t]
  \centering
  \includegraphics[width=\linewidth]{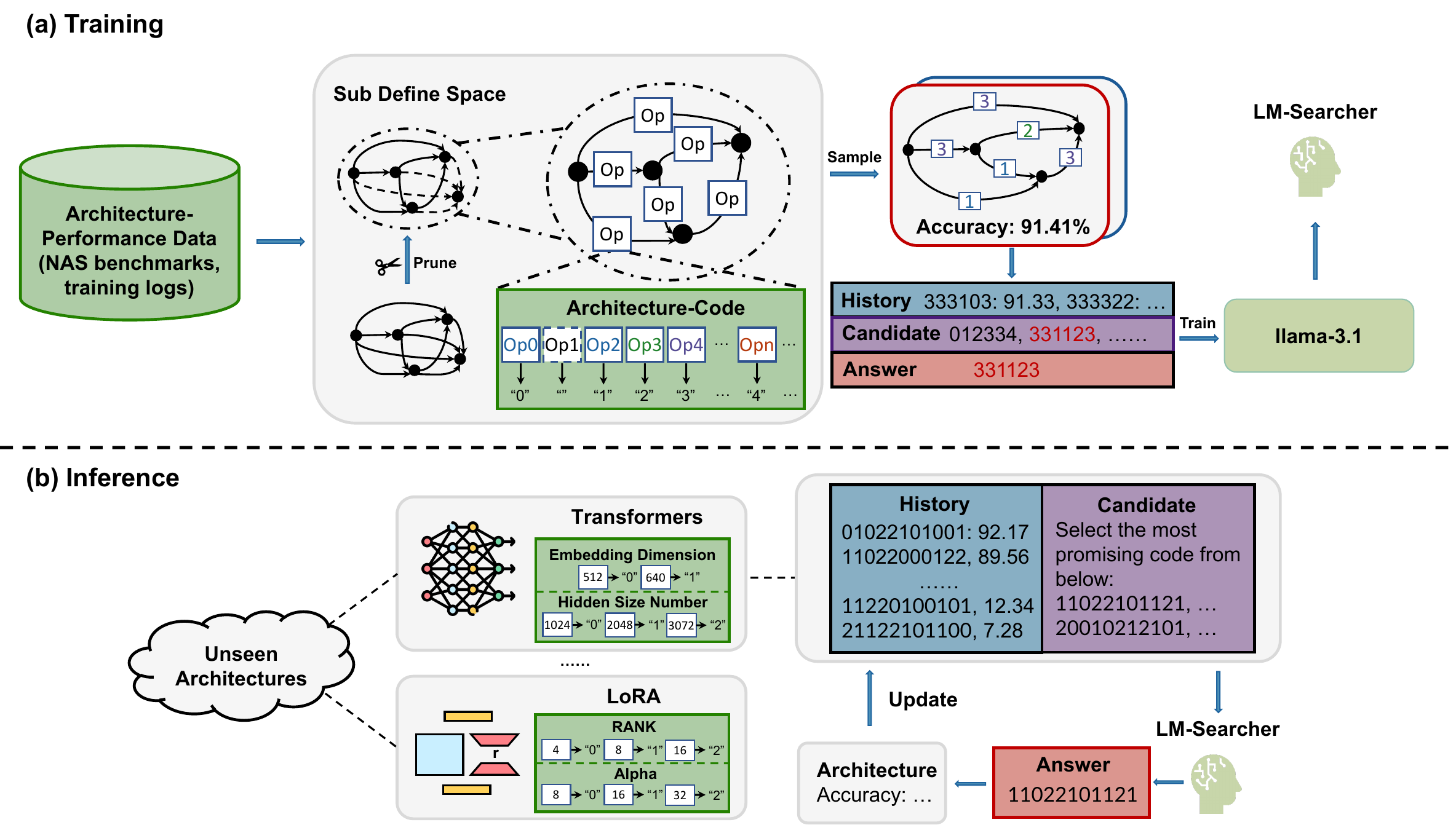}
  \caption{(a) \textbf{Training phase:} The full architecture space is pruned into diverse subspaces, from which we sample and encode architectures as numerical codes. Architecture-performance pairs form each training sample, with the top-performing code as the answer. Instructions include a "history" of pairs and a candidate set of codes for the LLM to select from.
(b) \textbf{Inference phase:} At inference, LM-Searcher frames optimization tasks as combinatorial problems using the same encoding, enabling unified and generalizable search across domains.}
  \label{fig:method_fig2}
\end{figure*}

\section{Method}
We introduce LM-Searcher, a task-agnostic neural architecture search framework powered by LLMs. By encoding cross-domain architectures into a unified representation, LM-Searcher leverages LLMs to iteratively analyze historical architecture-performance data, rank new candidates, and predict the most promising architectures until convergence. We begin by formally defining the problem to better leverage LLMs for NAS problems in Sec.~\ref{prob_formulation}. Then, we present the proposed LM-Searcher pipeline in Sec.~\ref{sec:lm-searcher}, which consists of three key components: 
Domain-agnostic Architecture Representation, Optimization Trajectory Data Curation, and Cross-domain Inference.

\subsection{Problem Formulation}
\label{prob_formulation}
Neural Architecture Search (NAS) involves finding an optimal neural network architecture from a predefined search space to maximize performance on specific tasks~\cite{zoph2016neural}. Formally, given a search space $S$ and history of evaluated architecture, the objective is to identify an architecture $a^*$ that maximizes the performance on the target task:

\begin{equation} \label{eq1}
    a^* = \underset{a \in S}{\text{argmax}} \; P(a)\text{,}
\end{equation}

\noindent where $P(a)$ denotes the target performance metric (e.g., accuracy or throughput). Utilizing LLMs with strong in-context learning and reasoning abilities, we aim at building a framework applicable to architecture search across domains, we model all NAS tasks in a consistent form by introducing a universal architecture encoding method and reformulating the architecture generation task as a ranking problem, which identify the optimal architecture from candidate architectures pool $C$, $C$ sampled from entire search space $S$. The optimization objective of Eq.~\ref{eq1} can be formulated into 

\begin{equation} \label{eq2}
    a^* = \underset{a \in C}{\mathop{\mathrm{argmax}}}\, P(a).
\end{equation}

This formulation simplifies the process of obtaining the optimal architecture $a^*$ from a subspace of candidates. In the next section, we introduce the details of training data curation for LLMs training and inference.

\subsection{LM-Searcher}
\label{sec:lm-searcher}

\textbf{Domain-agnostic Architecture Representation.} To enable domain-agnostic and LLM-friendly representation of architectures, we encode each architecture as a simplified numerical string called \textbf{NCode}. In this encoding, each digit corresponds to the index of a selected configuration option (such as an operator or a topology specification).  For instance, in NAS-Bench-201, an architecture is defined by assigning one of five operations—0 (zeroize), 1 (skip connection), 2 (1×1 convolution), 3 (3×3 convolution), or 4 (3×3 average pooling)—to each edge in a directed acyclic graph (DAG) consisting of six edges. Thus, every architecture can be uniquely represented by a six-digit code, where each digit (ranging from 0 to 4) denotes the operation on the corresponding edge (see Fig.~\ref{fig:intro_fig1}). Note that some operations, such as zeroize and skip connection, also influence the network topology. For example, an architecture with multiple 3×3 convolutions, a 1×1 convolution, and skip connections would be encoded as "333123", as illustrated in Fig.~\ref{fig:intro_fig1}. This approach ensures a compact and cross-domain model representation suitable for LLM processing, enabling the model to analyze relationships between different architectures based on their encoded architecture and corresponding performance.

\textbf{Optimization Trajectory Data Curation.} 
Leveraging our proposed NCode representation, we can unify and curate large-scale instruction-tuning data from existing NAS benchmarks~\cite{ying2019bench,zela2020surrogate} and training logs, enabling the training of a NAS LLM.

To construct high-quality training samples, we first sparsely prune the entire architecture search space, following~\cite{pham2018efficient}, to generate distinct subspaces for each training subnetwork. As shown in the left part of Fig.~\ref{fig:method_fig2} (a), each edge in the overall DAG is pruned with a 50\% probability, and each selectable configuration is also independently pruned with a 50\% probability. This independent pruning process produces diverse subnetworks, thus providing a sufficiently large and varied search space for constructing effective training data for LLMs.

In a pruned subspaces, we randomly sample 100-200 architectures to construct a training sample. However, since the search space is too large to exhaustively explore, retrieving the optimal history-prediction pair is infeasible. To address this, we reformulate the LLM’s task: instead of generating the best-performing architecture directly, it ranks a set of candidate architectures sampled from the pruned search space using predefined rules. As illustrated in Tab.~\ref{tab:prompt}, for each training example, we present the LLM with a history (a list of previous architectures and their accuracies), a set of candidate architectures, and ask it to select the most promising candidate based on patterns learned from the historical data. The training sample thus takes the form: "history (NCode-performance), candidates (NCode), answer (NCode)", where the answer is the actual best-performing architecture among the candidates.

By this means, we generate a dataset comprising 228k optimization trajectory, which are used to fine-tune open-source LLMs and obtain our final LLM-Searcher models.

\begin{table}[t]\fontsize{8.2}{2}\selectfont
\centering
\begin{tabularx}{\columnwidth}{X}
\toprule 
\textbf{Input} \\
\midrule
Please analyze the history, rank the candidate and output the highest-performing candidate. \\
\\
\\
\textbf{History}:\\ \\
\\
NCode: 03255564, accuracy: 94.28;\\
\\
NCode: 43212502, accuracy: 89.47;\\
\\
......\\
\\
NCode: 63421032, accuracy: 25.76;\\
\\
NCode: 53215432, accuracy: 14.13;\\ \\
\\
\textbf{Candidate}: \\
\\ \\
33513501 \\ 
\\
63225362 \\ \\ 
... \\ \\
41625214 \\ \\
\\

\midrule
\textbf{Output} \\
\midrule
63225362\\
\bottomrule 
\end{tabularx}
\caption{An example of the prompt. The LLM needs to reason and learn from the "history", rank the "candidate", and output which of the "candidate" is expected to perform best.}
\label{tab:prompt}
\end{table}

\textbf{Cross-domain Inference.} During inference, the NCode representation allows us to encode previously unseen architectures into a unified format suitable for processing by the LLM. As illustrated in Fig.~\ref{fig:method_fig2} (b), for example, when encoding a LoRA module with a search space of [4, 8, 16] for the embedding dimension, the choices 4, 8 and 16 are represented as 0, 1 and 2, respectively. 

To demonstrate cross-domain generalization, we evaluate LM-Searcher on both in-domain and out-of-domain tasks \textit{without specifically tuning}. Following the iterative approach used in NAS approaches~\cite{yuslimmable,zoph2016neural}, LM-Searcher samples an architecture (NCode) based on historical NCode-performance pairs and 10 randomly generated NCodes. The sampled architecture is then evaluated, and the historical NCode-performance pairs are updated for subsequent iterations.

\begin{table*}[t]
    \centering
    \resizebox{0.99\linewidth}{!}{
        \begin{tabular}{l|c|c|c|c|c|c|c|c|c|c}
          \toprule
          \multirow{3}{*}{Method} & \multicolumn{3}{c|}{\textbf{Classification (In-domain)}} & \multicolumn{2}{c|}{\textbf{Segmentation}} & \multicolumn{3}{c|}{\textbf{Generation}} & \textbf{Translation} & \textbf{Audio} \\
          \cmidrule(lr){2-4} \cmidrule(lr){5-6} \cmidrule(lr){7-9} \cmidrule(lr){10-10} \cmidrule(lr){11-11}
          & CIFAR10 & CIFAR100 & ImageNet & Kvasir & ISIC 2017 & \multicolumn{3}{c|}{Multi-Concept Customization} &  IWSLT’14 De-En & NB-ASR \\
          & Test Err.(\%)↓ & Top-1 Acc.(\%)↑ & Top-1 Acc.(\%)↑ & $S_\alpha$(\%)$\uparrow$ & Jac(\%)$\uparrow$ & CLIP-T↑ & CLIP-I↑ & DINO↑ & BLEU↑ & PER(\%)↓ \\
          \midrule
          DARTS \citep{liu2018darts} & 2.76 & 15.03 & 73.3 & - & - & - & - & - & - & - \\
          PC-DARTS \citep{xu2019pc} & \textbf{2.57} & - & 74.9 & - & - & - & - & - & - & - \\
          TE-NAS \citep{chen2021neural} & 2.63 & 71.24 & 73.8 & - & - & - & - & - & - & -\\
          AG-Net \citep{lukasik2022learning} & 5.63 & \textbf{73.51} & \textbf{76.5} & - & - & - & - & - & - & - \\
          DiNAS \citep{asthana2024multi} & 5.63 & \textbf{73.51} & 75.2 & - & - & - & - & - & - & -\\
          LLMatic~\cite{nasir2024llmatic} & 5.74 & 71.62 & - & - & - & - & - & - & - & - \\
          GENIUS~\cite{zheng2023can} & 6.21 & 70.91 & 74.9 & - & - & - & - & - & - & - \\
          \midrule
          Random Search~\cite{li2020random} & 3.41 & 72.07 & 72.56 & 91.71 & 76.28 & 0.654 & 0.728 & 0.385 & 33.00 & 21.50 \\
          Regularized Evolution~\cite{real2019regularized} & 3.34 & 71.81 & 74.5 & 91.89 & 77.18 & 0.659 & 0.727 & 0.396 & 33.45 & \textbf{21.40} \\
          \midrule
          \rowcolor[gray]{.9}
          \textbf{LM-Searcher} & \textbf{3.10} & \textbf{72.96} & \textbf{75.5} & \textbf{92.35} & \textbf{77.60} & \textbf{0.668} & \textbf{0.737} & \textbf{0.416} & \textbf{34.02} & 21.44 \\
          \bottomrule
        \end{tabular}
    }
    \caption{Results of LM-Searcher, compared to various traditional and LLM-based methods on in-domain image classification tasks and 4 out-of-domain tasks (\textit{Image Segmentation, Image Generation, Machine translation, Audio Recognition)}.}
    \label{tab:main_tab}
\end{table*}

\section{Experiment}

\subsection{LM-Searcher Training Details}
The supervised fine-tuning of LLaMA-3.1-8B on our curated dataset was performed with 320GB of total GPU memory (8×40GB) for 1 epoch. We utilize the codebase provided by~\cite{zheng2024llamafactory}\footnote{\url{https://github.com/hiyouga/LLaMA-Factory}} to fine-tune the LLaMA-3.1-8B model. The fine-tuning process employs an initial learning rate of $1 \times 10^{-5}$, a warm-up step ratio of 0.1, and a batch size of 8 with gradient accumulation steps set to 2. To ensure computational efficiency, we utilize mixed-precision training techniques. Additionally, we leverage DeepSpeed ZeRO-2~\cite{rajbhandari2020zero}\footnote{\url{https://github.com/microsoft/deepspeed}} to optimize memory usage and enhance scalability during the large-scale training process.

\subsection{LM-Searcher for Diverse Tasks}
We apply LM-Searcher to perform neural architecture search across diverse domains. For \textbf{in-domain} image classification tasks, we evaluate on CIFAR-10, CIFAR-100, and ImageNet-1K~\cite{deng2009imagenet}\footnote{\url{https://www.image-net.org/}}. Please refer to Appendix~\ref{appendix:in_domain} for more details. For \textbf{out-of-domain} tasks, in the \textit{image segmentation} task, we use the Kvasir-SEG~\cite{jha2020kvasir} and ISIC2017~\cite{codella2018skin} datasets. For \textit{image generation}, we explore parameter-efficient fine-tuning (PEFT) of diffusion models~\cite{rombach2022high}. Additionally, we explore efficient transformer architectures for \textit{machine translation}~\cite{wang2020hat} and \textit{audio recognition} using NAS-Bench-ASR~\cite{mehrotra2021bench}.

The main results are shown in Tab.~\ref{tab:main_tab}.  LM-Searcher exhibits strong search capabilities across diverse domains, including in-domain tasks like CIFAR10 image classification and out-of-domain tasks such as audio recognition, searching efficient transformer architecture for machine translation, and LoRA-based image generation. Compared to other NAS methods, including LLM-based methods, our LM-searcher can not only achieve comparable performance in-domain tasks, but also directly applied to out-of-domain tasks. Additional implementation details and experimental results can be found in Sec.~\ref{experiment:out_domain}.

We re-implemented the random search and regularized evolution algorithms to compare their performance with our method on out-of-domain tasks. LM-Searcher consistently outperforms evolutionary algorithms (EA) and random search in most scenarios. For in-domain tasks, it achieves performance comparable to specialized NAS methods, while maintaining broader applicability.

\subsection{Out-of-Domain Tasks Experimental Details}
\label{experiment:out_domain}

\textbf{Image Segmentation.} Segment Anything (SAM)~\cite{kirillov2023segment} is a foundation model for image segmentation. We employ LM-Searcher to explore the parameter-efficient-finetuning (PEFT) architecture of SAM for image segmentation tasks. We utilize an architecture search space following Conv-LoRA~\cite{zhongconvolution}\footnote{\url{https://github.com/autogluon/autogluon/tree/master/examples/automm/Conv-LoRA}}, which integrates convolution operations and LoRA modules into SAM's ViT encoder~\cite{dosovitskiy2020image}. We employ LM-Searcher to determine effective rank values of 32 LoRA modules from [3, 6, 12, 24]. As rank value of each LoRA module can be configured independently, the total number of unique architectures reaches up to $10^{19}$. 

During the search phase, we use a proxy task to train and evaluate sampled models on the Kvasir-SEG polyp dataset~\cite{jha2020kvasir} and the skin lesion segmentation dataset~\cite{codella2019skin}. To ensure time efficiency, training is early stopped after one epoch. The performance of each architecture is assessed using a weighted sum of validation metrics. We conduct the search over 200 iterations and retrain the top-performing model for 30 epochs to evaluate its final performance.

To comprehensively evaluate the effectiveness of the searched architecture across diverse domains, we conduct extensive experiments on multiple downstream tasks. In the medical imaging domain, we validate the architecture's performance by fine-tuning the SAM model on three benchmark datasets: the Kvasir-SEG polyp dataset~\cite{jha2020kvasir} for gastrointestinal polyp segmentation, CVC-612~\cite{codella2019skin} for colorectal lesion analysis, and ISIC2017~\cite{codella2018skin} for skin lesion classification. These experiments systematically demonstrate the architecture's capability in medical domain adaptation and fine-tuning scenarios. For natural image analysis, we assess the model's performance using CAMO~\cite{le2019anabranch} and SBU~\cite{vicente2016large}. Furthermore, we extend our evaluation to additional application domains, including agricultural analysis through leaf disease segmentation and remote sensing through road segmentation, thereby demonstrating the searched architecture's versatility across different fields.

\begin{table*}[t]
\centering
\resizebox{0.99\linewidth}{!}{
    \renewcommand{\arraystretch}{1.2}
    \setlength{\tabcolsep}{4pt}
    \begin{tabular}{lccccccccccc}
    \toprule
    \textbf{Method} & \multicolumn{2}{c}{\textbf{Kvasir}} & \multicolumn{2}{c}{\textbf{CVC-612}} & \textbf{ISIC 2017} & \multicolumn{3}{c}{\textbf{CAMO}} & \textbf{SBU} & \textbf{Leaf} & \textbf{Road} \\
                    & $S_\alpha\uparrow$ & $\quad E_\phi\uparrow$ & $S_\alpha\uparrow$ & $E_\phi\uparrow$ & Jac$\uparrow$ & $S_\alpha\uparrow$ & $E_\phi\uparrow$ & $F_\beta^\omega\uparrow$ & BER$\downarrow$ & IoU$\uparrow$ & IoU$\uparrow$ \\
    \midrule
    BitFit~\cite{zaken2021bitfit}          & 90.8 $\pm$ 0.57 & 93.8 $\pm$ 0.98 & 89.0 $\pm$ 0.40 & 91.6 $\pm$ 0.98 & 76.4 $\pm$ 0.45 & 86.8 $\pm$ 0.33 & 90.7 $\pm$ 0.28 & 81.5 $\pm$ 0.19 & 3.2 $\pm$ 0.13 & 71.4 $\pm$ 1.15 & 60.6 $\pm$ 0.15 \\
    Adapter~\cite{houlsby2019parameter}         & 91.2 $\pm$ 0.23 & 94.0 $\pm$ 0.16 & 89.3 $\pm$ 0.43 & 92.0 $\pm$ 0.63 & 76.7 $\pm$ 0.66 & 87.7 $\pm$ 0.10 & 91.3 $\pm$ 0.40 & 82.8 $\pm$ 0.35 & 2.8 $\pm$ 0.09 & 72.1 $\pm$ 0.47 & 61.5 $\pm$ 0.11 \\
    VPT~\cite{liu2024visual}             & 91.5 $\pm$ 0.23 & 94.3 $\pm$ 0.06 & 91.0 $\pm$ 0.94 & \textbf{93.7 $\pm$ 1.41} & 76.9 $\pm$ 0.94 & 87.4 $\pm$ 0.60 & 91.4 $\pm$ 0.68 & 82.1 $\pm$ 0.75 &  2.7 $\pm$ 0.06 & 73.6 $\pm$ 0.26 & 60.2 $\pm$ 1.87 \\
    LST~\cite{sung2022lst}             & 89.7 $\pm$ 0.25 & 93.3 $\pm$ 0.37 & 89.4 $\pm$ 0.37 & 92.4 $\pm$ 0.54 & 76.4 $\pm$ 1.05 & 83.3 $\pm$ 0.28 & 88.0 $\pm$ 0.23 & 77.1 $\pm$ 0.02 & 3.2 $\pm$ 0.01 & 70.2 $\pm$ 0.87 & 60.2 $\pm$ 0.26 \\
    SAM-Adapter~\cite{chen2023sam}     & 89.6 $\pm$ 0.24 & 92.5 $\pm$ 0.10 & 89.6 $\pm$ 0.22 & 92.4 $\pm$ 1.06 & 76.1 $\pm$ 0.45 & 85.6 $\pm$ 0.26 & 89.6 $\pm$ 0.55 & 79.8 $\pm$ 0.89 & 3.1 $\pm$ 0.06 & 71.4 $\pm$ 0.20 & 60.6 $\pm$ 0.06 \\
    SSF~\cite{lian2022scaling}             & 91.3 $\pm$ 0.87 & 93.9 $\pm$ 1.49 & 89.6 $\pm$ 0.37 & 91.9 $\pm$ 0.79 & 76.6 $\pm$ 0.19 & 87.5 $\pm$ 0.11 & 91.4 $\pm$ 0.16 & 82.6 $\pm$ 0.12 & 3.2 $\pm$ 0.05 & 71.5 $\pm$ 0.63 & 61.6 $\pm$ 0.03 \\
    LoRA~\cite{hu2021lora}            & 91.2 $\pm$ 0.28 & 93.8 $\pm$ 0.22 & 90.7 $\pm$ 0.04 & 92.5 $\pm$ 0.41 & 76.6 $\pm$ 0.23 & 88.0 $\pm$ 0.24 & 91.9 $\pm$ 0.42 & 82.8 $\pm$ 0.16 & 2.7 $\pm$ 0.08 & 73.7 $\pm$ 0.20 & 62.2 $\pm$ 0.21 \\
    Conv-LoRA~\cite{zhongconvolution}       & 91.9 $\pm$ 0.64 & 94.3 $\pm$ 0.77 & 90.1 $\pm$ 0.31 & 92.3 $\pm$ 0.42 & 77.0 $\pm$ 0.51 & 88.6 $\pm$ 0.12 & 92.4 $\pm$ 0.17 & 83.1 $\pm$ 0.19 & 2.8 $\pm$ 0.02 & 73.9 $\pm$ 0.20 & 62.1 $\pm$ 0.23 \\
    \midrule
    \rowcolor[gray]{.9} 
    \textbf{LM-Searcher} & \textbf{92.4 $\pm$ 0.43} & \textbf{94.7 $\pm$ 0.47} & \textbf{91.1 $\pm$ 0.47} & 93.2 $\pm$ 0.70 & \textbf{77.6 $\pm$ 0.40} & \textbf{88.9 $\pm$ 0.04} & \textbf{92.7 $\pm$ 0.15} & \textbf{83.9 $\pm$ 0.32} & \textbf{2.6 $\pm$ 0.04} & \textbf{74.2 $\pm$ 0.28} & \textbf{62.6 $\pm$ 0.04} \\
    \bottomrule
    \end{tabular}
}
\caption{Performance comparison of various PEFT methods across multiple datasets for segmentation tasks. Results are reported as average values with standard errors, calculated over five experimental runs. Notably, LM-Searcher performs NAS in a zero-shot manner without finetuning on the target domain.}
\label{tab:seg}
\end{table*}

Tab.~\ref{tab:seg} compares the LoRA architecture searched using LM-Searcher with other PEFT models on several segmentation tasks. To account for training variability due to random initialization, each sampled architecture is trained five times, and the average performance is reported. Our LM-Searcher consistently outperforms the Conv-LoRA baseline, achieving superior performance across most evaluation metrics compared to other PEFT approaches. These results indicate that LM-Searcher exhibits strong generalization capabilities across datasets from different domains.

\textbf{Image Generation.} We employs LM-Searcher to identify optimal LoRA configurations for integrating new concepts into a pre-trained Stable Diffusion model~\cite{rombach2022high} following the Mix-of-Show~\cite{gu2023mix}\footnote{\url{https://github.com/TencentARC/Mix-of-Show}} setting.We partition the LoRA modules in the Diffusion UNet into 48 distinct groups and explore adaptation ranks for each group from the set [4, 8, 16]. In the search phase, LM-Searcher is utilized to sample a model which is subsequently finetuned on 15 images related with a concept. Upon completion of 1,000 training iterations, the model's performance is evaluated using 11 validation prompts, with each prompt generating 8 images for assessment. To quantify the model's capabilities, we utilize the CLIP-Score~\cite{hessel2021clipscore} and the CLIP image alignment score as proxy metrics. These metrics effectively measure the model's caption-following capability and its ability to preserve identity, respectively. The overall performance of an architecture is computed as a weighted sum of two metrics. Upon concluding the search stage, we select the architecture with the highest performance score and proceed to fine-tune plug-and-play LoRA models individually across several concepts. The LoRA models are fused into one SD-V1.5 model with Gradient Fusion~\cite{gu2023mix} technique. After fusion, the Stable Diffusion model is tested using 20 prompts per concept, with each prompt generating 50 images. 

\begin{table}[t]
\centering
\resizebox{0.99\linewidth}{!}{
\begin{tabular}{l|ccc}
\toprule
\textbf{Method} & \textbf{CLIP-T} & \textbf{CLIP-I} & \textbf{DINO} \\ 
\midrule
P+~\cite{voynov2023p+}               & 0.686           & 0.670           & 0.372               \\ 
Custom Diffusion~\cite{kumari2023multi} & 0.650           & 0.694           & 0.379            \\ 
LoRA~\cite{hu2021lora}             & \textbf{0.700}           & 0.555           & 0.359           \\
ED-LoRA~\cite{gu2023mix}      & 0.662           & 0.731           & 0.404           \\ 
\midrule
    \rowcolor[gray]{.9} 
\textbf{LM-Searcher}      & 0.668           & \textbf{0.737}           & \textbf{0.416}           \\ 
\bottomrule
\end{tabular}
}
\caption{Performance Comparison of PEFT Methods in Multi-Concept Customization for Stable Diffusion Models. Evaluating CLIP-T (Text), CLIP-I (Image), and DINO Metrics. LM-Searcher conducts LoRA configurations search without additional training for image generation tasks.}
\label{tab:lora_diffusion}
\end{table}

We report the CLIP text similarity, CLIP image similarity and DINO image similarity of our approach and other PEFT methods in Tab.~\ref{tab:lora_diffusion}. The model architecture identified by LM-Searcher exhibits superior performance compared to state-of-the-art ED-LoRA across both text similarity and image similarity metrics. These findings indicate that LM-Searcher effectively discovers optimal architectures that not only preserve custom concept identity but also maintain strong image-text alignment. While our model achieves a slightly lower CLIP text score than both P+~\cite{voynov2023p+} and LoRA, it significantly surpasses these approaches in image alignment capability, outperforming P+ by 6.7\% and LoRA by 28.2\% in CLIP-I.

\textbf{Efficient Transformer for Machine Translation.} To further demonstrate the cross-domain generalizability of LM-Searcher, we apply it to explore efficient transformer architectures for machine translation on edge devices following the HAT~\cite{wang2020hat} setting, with a search space encompassing embedding dimensions from [512, 640], hidden dimensions from [1024, 2048, 3072], attention head numbers from [4, 8], and decoder layer numbers from [1, 2, 3, 4, 5, 6]. After training a super-transformer, we sample and evaluate 3K sub-transformers based on their validation losses on the IWSLT'14 De-En validation set under a GPU latency constraint of 500 ms, selecting the sub-transformer with the lowest validation loss for retraining. The retrained model is then evaluated using BLEU and SacreBLEU scores~\cite{post2018call}.

\textbf{Audio Recognition.} To evaluate LM-Searcher's effectiveness in the audio domain. We utilize the tabular NAS-Bench-ASR~\cite{mehrotra2021bench} to explore architectures for audio recognition. The search space is modeled by a DAG using four nodes, with each node's configuration consists of a primary operation (e.g., linear operations, convolution operations with various kernel sizes and dilation rates, or a zero operation) and a skip connection operation (e.g., identity or zeroize).

\begin{table}[t]
    \centering
    \resizebox{0.99\linewidth}{!}{
    \begin{tabular}{l|cccc}
        \toprule
        \textbf{Model} &\textbf{Base Model} & \textbf{NAS-Bench-101} & \textbf{CIFAR100} & \textbf{ImageNet} \\
        \midrule
        LM-Searcher-1B &{LLaMA-3.1-1B}& 93.85 & 72.31 & 73.52 \\
        LM-Searcher-3B & {LLaMA-3.1-3B} & 93.85 & 72.60 & 73.93 \\
        \rowcolor[gray]{.9} 
        LM-Searcher-8B & {LLaMA-3.1-8B} & \textbf{93.89} & \textbf{72.96} & \textbf{75.5} \\
        \bottomrule
    \end{tabular}
    }
    \caption{Performance metrics for LM-Searcher with different base model size.}
    \label{tab:ablation_model_size}
\end{table}

\section{Ablation Studies}
In this section, we conduct ablation studies to analyze the impact of key components in our LM-Searcher framework. Specifically, we evaluate how our proposed pruning-based subspace sampling and task reformulation affect the overall performance in Sec.~\ref{stra}. We also study the impact of the LLM sizes in Sec.~\ref{llmsize} and whether shuffling the mappings between architectures and performances affects the results in Sec.~\ref{mapping}.

\begin{table}[t]
    \centering
    \resizebox{1.0\linewidth}{!}{
        \begin{tabular}{l|c|c|c|c|c|c}
          \toprule
          \multirow{2}{*}{\textbf{Mapping}} & \multicolumn{2}{c|}{\textbf{CIFAR10}} & \multicolumn{2}{c|}{\textbf{CIFAR100}} & \multicolumn{2}{c}{ \textbf{ImageNet16-120}
} \\
 & Validation & Test & Validation & Test & Validation & Test \\
          \midrule
          Shuffle & 91.22 & 93.87 & 72.20 & 71.89 & 45.93 & 46.25 \\       
          \rowcolor[gray]{.9} 
          w/o Shuffle & \textbf{91.52} & \textbf{94.20} & \textbf{72.82} & \textbf{72.96} & \textbf{46.48} & \textbf{46.51} \\
          \bottomrule
        \end{tabular}
    }
    \caption{Ablation on whether to shuffle the mapping between architectures and performance.}
    \label{tab:ablation_nas_bench_201_control}
\end{table}

\subsection{Effects of Training Strategy}
\label{stra}
We assess the effects of our proposed training strategy by evaluating LLaMA-3.1-8B, fine-tuned using datasets constructed with different settings.
Specifically, we compare our full training strategy (denoted as "Ours") with two ablation variants: (1) w/o pruning-based subspace sampling, where all the samples in the training data are sampled from the entire search space instead of from a pruned sub search space, and (2) w/o task reformulation, where the LLM is prompted to directly generate better architectures instead of ranking from the candidates.
As shown in Tab.~\ref{tab:ablation_nas_bench_201_train}, our task reformulation technique leads to significant performance improvements on the CIFAR10, CIFAR100, and ImageNet16-120 datasets. Additionally, the proposed pruning-based sampling technique further enhances the LLM’s ability to search for neural architectures more effectively.

\subsection{Impacts of LLM Size}
\label{llmsize}
To investigate the impact of LLM size on performance, we evaluate LM-Searcher with different model scales, with results shown in Tab.~\ref{tab:ablation_model_size}. We train models at different scales and report test accuracy on NAS-Bench-101, CIFAR-100, and ImageNet.
Our results show a consistent improvement as the LLM size increases, indicating that larger models can better capture the promising architecture patterns.

\subsection{Architecture-performance Mapping}
\label{mapping}

To further validate LM-Searcher's capability in identifying efficient architectures through pattern analysis of historical experimental data, we conducted an ablation study where we intentionally randomized the architecture-performance mapping in NAS-Bench-201. This manipulation introduces noise into the information provided to the LLM. As depicted in Tab.~\ref{tab:ablation_nas_bench_201_control}, the performance degrades significantly when the architecture-performance pairs provided to LM-Searcher are corrupted by noise. This result shows that LM-Searcher is able to utilize Ncode-performance pairs provided in the context to effectively search architectures. More experimental analysis are provied in Appendix~\ref{appendix:analysis}.

\subsection{Trade-off Between Specialization and Generalization}
We conducted in-domain fine-tuning with LM-Searcher by encoding architectures with more domain-specific information rather than simple digit encoding. For example, a structure originally encoded as \texttt{333123} is now encoded as \texttt{\seqsplit{|conv3X3\textasciitilde0|+|conv3X3\textasciitilde0|+|conv3X3\textasciitilde1|+|skip\_connect\textasciitilde0|+|conv1X1\textasciitilde1|+|conv3X3\textasciitilde2|}}, where the numbers following the tilde (\textasciitilde) indicate which previous feature (node) the operator (edge in the DAG) connects to. As shown in Tab.~\ref{tab:ablation_trade_off}, after fine-tuning on 1000 samples created using this encoding method, LM-Searcher achieves performance that approaches or even surpasses existing state-of-the-art methods. However, finetuning the model with such domain-specific samples restricts cross-domain generalization, resulting in performance degradation in other domains (0.97\% decline on IWSLT'14 De-En machine translation and 0.13\% decline on NAS-Bench-ASR). 

\begin{table}[t]
    \centering
    \resizebox{1.0\linewidth}{!}{
        \begin{tabular}{l|c|c|c|c|c|c}
          \toprule
          \multirow{2}{*}{\textbf{Config}} & \multicolumn{2}{c|}{\textbf{CIFAR10}} & \multicolumn{2}{c|}{\textbf{CIFAR100}} & \multicolumn{2}{c}{ \textbf{ImageNet16-120}
} \\
 & Validation & Test & Validation & Test & Validation & Test \\
          \midrule
          w/o pruning-based sampling & 90.83 & 93.39 & 71.08 & 70.98 & 44.83 & 43.98 \\       
          w/o task reformulation & 89.10 & 92.16 & 69.33 & 69.51 & 43.38 & 42.53 \\
          \midrule
          \rowcolor[gray]{.9} 
          \textbf{Ours} & \textbf{91.52} & \textbf{94.20} & \textbf{72.82} & \textbf{72.96} & \textbf{46.48} & \textbf{46.51} \\
          \bottomrule
        \end{tabular}
    }
    \caption{Performance on NAS-Bench-201 using different training strategy, we report the average accuracy of 5 runs.}
    \label{tab:ablation_nas_bench_201_train}
\end{table}

\begin{table}[t]
    \centering
    \resizebox{1.0\linewidth}{!}{
        \begin{tabular}{l|ccccc}
        \toprule
        Method & \textbf{CIFAR10}$\uparrow$ & \textbf{CIFAR100}$\uparrow$ & \textbf{ImageNet-16}$\uparrow$ & \textbf{IWSLT’14De-En}$\uparrow$ & \textbf{NB-ASR}$\downarrow$  \\ 
        \midrule
        DARTS~\citep{liu2018darts} & 54.30 & 39.77 & 16.32 & - & - \\
        SGNAS~\citep{huang2021searching} & 93.53 & 70.31 & 44.98 & - & - \\
        DiNAS~\citep{asthana2024multi} & \textbf{94.37} & \textbf{73.51} & 45.51 & - & - \\
        AG-Net~\citep{lukasik2022learning} & \textbf{94.37} & \textbf{73.51} & 46.42 & - & - \\
        \midrule
        LM-Searcher-FT &  94.36 & \textbf{73.51} & \textbf{46.54} & 33.05 & 21.57 \\
        \rowcolor[gray]{.9} 
        LM-Searcher & 94.20 & 72.96 & 46.51 & \textbf{34.02} & \textbf{21.44} \\
        \bottomrule
        \end{tabular}
    }
    \caption{Performance on in-domain image classification (CIFAR-10, CIFAR-100, ImageNet-16 from NAS-Bench-201) and out-of-domain tasks (IWSLT’14 De–En machine translation, NB-ASR speech recognition). LM-Searcher-FT is fine-tuned on 1,000 domain-specific samples.}
    \label{tab:ablation_trade_off}
\end{table}

\section{Conclusion}
In this work, we introduced LM-Searcher, a general-purpose framework for neural architecture optimization that leverages LLMs and a novel cross-domain numerical encoding, NCode. By reformulating neural architecture search as a ranking problem and employing a pruning-based subspace sampling strategy for data curation, LM-Searcher demonstrates robust performance across both in-domain and out-of-domain tasks. Our results highlight the potential of LLMs to serve as flexible, domain-agnostic architecture search models, reducing reliance on domain-specific expertise and manual tuning. We believe our approach sheds new insights into scalable and adaptable methods in automated model search and design. For future work, we aim to expand the training data to encompass broader search spaces and enhance LM-Searcher’s efficiency during inference.

\section{Acknowledgments}
This project is funded in part by National Key R\&D Program of China Project 2022ZD0161100, by the Centre for Perceptual and Interactive Intelligence (CPII) Ltd under the Innovation and Technology Commission (ITC)’s InnoHK, and in part by NSFC-RGC Project N\_CUHK498/24. 

\section{Limitations}

While LM-Searcher demonstrates promising generality and flexibility across a range of neural architecture search domains, several limitations remain. First, our current focus has been on validating the cross-domain capabilities and generalization of the search model, rather than achieving state-of-the-art (SOTA) performance on specific benchmarks. As a result, there is still a performance gap between LM-Searcher and highly specialized NAS methods tailored for individual domains. Additionally, our approach relies on the availability and quality of existing architecture-performance metadata, which may limit its applicability to domains lacking comprehensive benchmarks. Finally, the NCode representation, while effective for unifying diverse tasks, may not capture all the nuanced architectural details needed for highly specialized designs. We plan to address these limitations in future work by optimizing the model for SOTA performance, expanding to new domains, and further refining the architecture encoding scheme.

\bibliography{custom}

\appendix
\label{sec:appendix}

\section{In-Domain Tasks Experimental Details}
\label{appendix:in_domain}
\paragraph{NAS-Bench-201}
NAS-Bench-201~\footnote{\url{https://github.com/D-X-Y/NAS-Bench-201}}~\cite{dong2020bench} is a widely used benchmark that provides pre-computed results of 15625 architectures pretrained and evaluated on three different image classification datasets.

\paragraph{ImageNet-1K Dataset}
The ImageNet-1K~\footnote{\url{https://www.image-net.org/}} dataset is a widely used benchmark in computer vision, particularly for image classification tasks. It comprises images from 1,000 distinct object categories. The training set contains approximately 1.2 million labeled images, while the validation set includes 50,000 images, with 50 samples per class.

We evaluate LM-Searcher's effectiveness on searching architectures for image classification under the large-scale DARTS search space. On the CIFAR-10 dataset, we employ an efficient proxy task—training each sampled architecture for only 10 epochs with a cosine learning rate scheduler. During the search phase, LM-Searcher samples 500 unique architectures. We then fully retrain the top 5 models for 200 epochs on CIFAR-10 training set to identify the highest performing architecture. On the ImageNet dataset, we utilize LM-Searcher to sample a total of 1K architectures from the surrogate benchmark~\cite{zela2020surrogate}. We retrain the top performing model on ImageNet training set for 250 epochs. Both training on CIFAR-10 and ImageNet use the SGD optimizer with hyperparameters following~\cite{liu2018darts}. Test set accuracy is reported and compared with other NAS methods that search architectures on CIFAR-10 and retrain on ImageNet. For a fair comparison, we re-implement GENIUS on CIFAR-10 using the official codebase\footnote{\url{https://github.com/mingkai-zheng/GENIUS}}.

To benchmark LM-Searcher's performance against other NAS approaches. We utilize the widely-used NAS-Bench-201~\cite{dong2020bench}. To mitigate the effects of randomness, each search experiment is conducted over five independent runs. As depicted in Table~\ref{tab:nas_bench_201}, LM-Searcher achieves competitive performance across all the dataset.

\section{Metric Explanation}

\paragraph{S-measure ($S_\alpha$)} S-measure evaluates the structural similarity between the predicted segmentation and ground truth. It considers both region-aware ($S_r$) and object-aware ($S_o$) structural similarities:

$$S_\alpha = \alpha \cdot S_o + (1-\alpha) \cdot S_r$$

where $\alpha$ is typically set to 0.5. The object-aware similarity focuses on the foreground object consistency, while region-aware similarity evaluates the overall structural information.

\paragraph{E-measure ($E_\phi$)} The Enhanced-alignment measure captures both pixel-level matching and image-level statistics. It is computed as:

$$E_\phi = \frac{1}{W \times H} \sum_{x=1}^{W} \sum_{y=1}^{H} \phi_{FM}(x,y)$$

where $\phi_{FM}$ is the enhanced alignment matrix, and $W$, $H$ are the width and height of the image.

\paragraph{Jaccard Index (Jac)} Jaccard Index is also known as Intersection over Union (IoU), it measures the overlap between predicted and ground truth regions:

$$\text{Jaccard} = \frac{|P \cap G|}{|P \cup G|}$$

where $P$ is the predicted segmentation and $G$ is the ground truth.

\paragraph{F-measure ($F_\beta^\omega$)} The weighted F-measure combines precision and recall with adaptive weights:

$$F_\beta^\omega = \frac{(1+\beta^2) \cdot \text{Precision}^\omega \cdot \text{Recall}^\omega}{\beta^2 \cdot \text{Precision}^\omega + \text{Recall}^\omega}$$

where the superscript $\omega$ indicates weighted versions that emphasize errors in important regions.

\paragraph{Boundary Error Rate (BER)} The boundary error rate specifically evaluates the accuracy of boundary detection, calculated as:

$$\text{BER} = \frac{|B_P \setminus B_G| + |B_G \setminus B_P|}{|B_G|}$$

where $B_P$ and $B_G$ are the predicted and ground truth boundaries, respectively.

\paragraph{Bilingual Evaluation Understudy (BLEU)} Bilingual Evaluation Understudy (BLEU) is an automatic metric for evaluating machine translation and other text generation tasks by measuring how closely a system’s output matches one or more reference translations. It computes modified n-gram precision: counts of overlapping 1- to 4-grams between the candidate and references, clipped so repeated n-grams aren’t over-rewarded. These precisions are combined via a geometric mean and multiplied by a brevity penalty that downweights outputs shorter than the references.

\paragraph{Phoneme Error Rate (PER).}
Phoneme Error Rate (PER) is a standard metric for evaluating phoneme-level recognition or alignment systems. Given a reference phoneme sequence and a hypothesized sequence, PER is computed as the edit distance between the two sequences—decomposed into substitutions (S), deletions (D), and insertions (I)—normalized by the length of the reference sequence (N). Formally,
\[
\mathrm{PER} = \frac{S + D + I}{N} \times 100\%,
\]
where $S$, $D$, and $I$ are obtained via an optimal alignment (e.g., dynamic programming for Levenshtein distance).

\section{Experimental Analysis}
\label{appendix:analysis}
\subsection{LM-Searcher Search behavior}
\begin{figure}[htbp]
    \centering
    \subfigure[CIFAR10 dataset]{
        \includegraphics[width=0.5\textwidth]{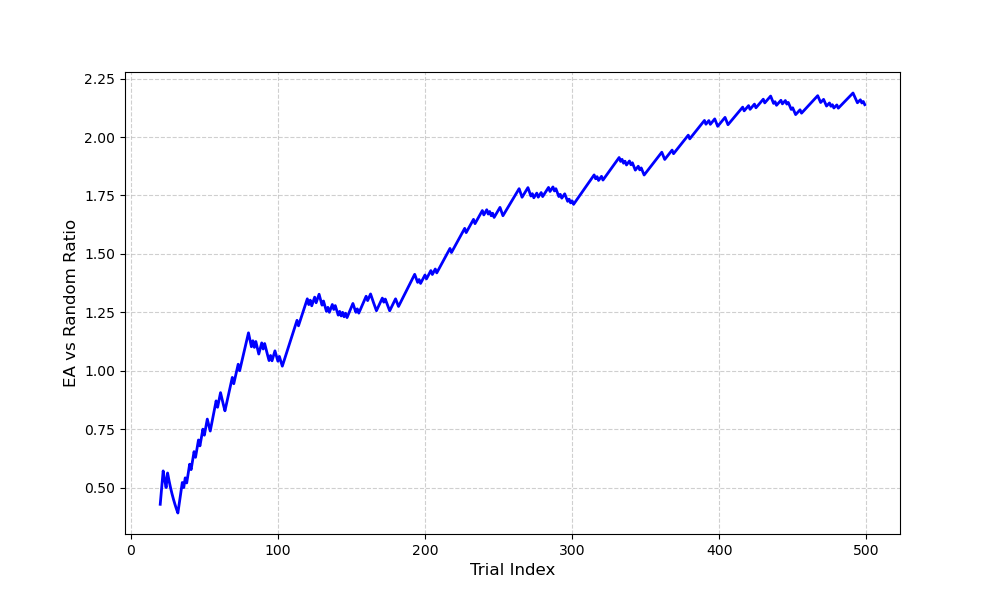}
    }
    \subfigure[CIFAR100 dataset]{
        \includegraphics[width=0.5\textwidth]{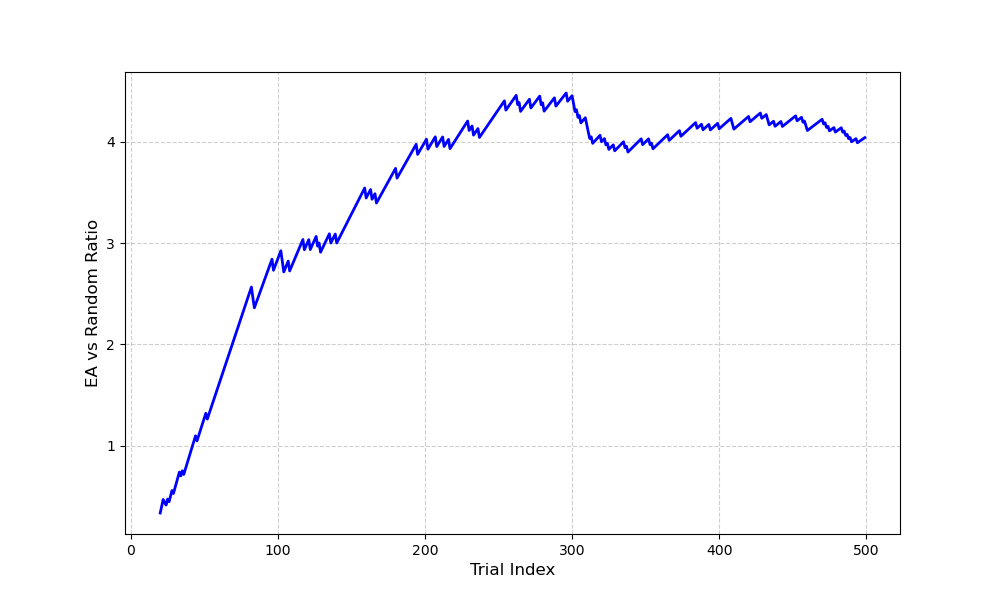}
    }
    \subfigure[ImageNet-16 dataset]{
        \includegraphics[width=0.5\textwidth]{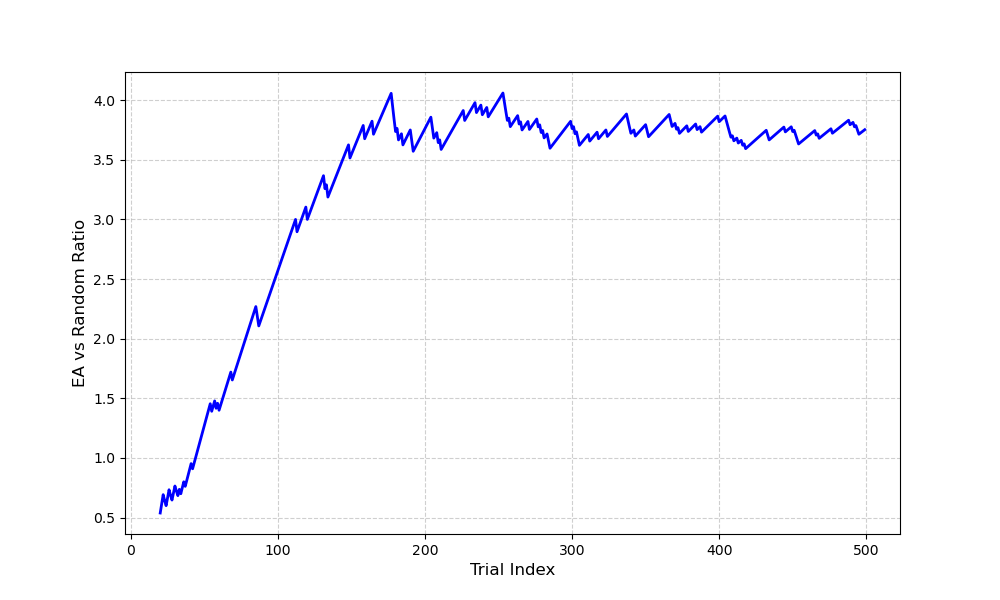}
    }
    \caption{}
    \label{fig:ratio_curve}
\end{figure}

In this section, we investigate the selection behavior of LM-Searcher by analyzing which architectures it tends to prefer. To this end, we generate half of the candidate architectures using a random search algorithm and the other half using evolutionary algorithms (EA). During the search phase, we track whether LM-Searcher selects candidate architectures originating from random search or EA. Specifically, we report the ratio of random search selections to EA selections as a function of the trial iteration number. As depicted in Fig.~\ref{fig:ratio_curve}, in the earlier stage of searching, LM-Searcher choose more candidates generated by random search than EA. As the search continues to iterate, the proportion of EA selection grows up with the trial iteration number, and surpassing the portion of random search selection after approximately 50-100 trials. A similar trend is observed when searching on the CIFAR-10, CIFAR-100, and ImageNet-16 datasets.

\subsection{LM-Searcher Attention Pattern}
We visualize the attention maps produced by LM-Searcher during the search process on NAS-Bench-201. Table~\ref{tab:attn_coe} reports the Pearson correlation coefficients between attention scores (measured between candidate NCode tokens and history NCode tokens) and two specific attributes of the history tokens: their performance and their similarity to the candidate NCode. As shown in the table, the attention scores exhibit a higher Pearson correlation with the performance of history NCode tokens than with their string similarity to the candidate NCode. This indicates that LM-Searcher does not solely depend on the external similarity to predict architectures. Notably, the negative correlation between attention scores and performance indicates that architectures with lower performance may provide more valuable information for efficient exploring the architecture search space.

\begin{table}[t]
    \centering
    \resizebox{0.95\linewidth}{!}{
        \begin{tabular}{l|c|c|c|c|c|c}
          \toprule
          \multirow{2}{*}{\textbf{Method}} & \multicolumn{2}{c|}{\textbf{CIFAR10}} & \multicolumn{2}{c|}{\textbf{CIFAR100}} & \multicolumn{2}{c}{ \textbf{ImageNet16-120}} \\

          & Validation & Test & Validation & Test & Validation & Test  \\
          \midrule
          
          
          DARTS \citep{liu2018darts} & 39.77 & 54.30 & 15.03 & 15.61 & 18.87 & 16.32 \\
          SGNAS \citep{huang2021searching} & 90.18 & 93.53 & 70.28 & 70.31 & 44.65 & 44.98 \\
          AG-Net \citep{lukasik2022learning} & 91.60 & \textbf{94.37} & \textbf{73.49} & \textbf{73.51} & \textbf{46.73} & \textbf{46.42} \\
          DiNAS \citep{asthana2024multi} & \textbf{91.61} & \textbf{94.37} & \textbf{73.49} & \textbf{73.51} & 46.66 & 45.41 \\
          \midrule
          
          \rowcolor{gray!10}
          \multicolumn{7}{l}{\textit{LLM-based}} \\
          GENIUS \citep{zheng2023can} & 91.07 & 93.79 & 70.96 & 70.91 &  45.29 & 44.96  \\
          LLMatic \citep{nasir2024llmatic} & - & \textbf{94.26} & - & 71.62 &  - & 45.87  \\
          \midrule
          \rowcolor[gray]{.9}
          \textbf{LM-Searcher} & \textbf{91.52} & 94.20 & \textbf{72.82} & \textbf{72.96} & \textbf{46.48} & \textbf{46.51}  \\
          \bottomrule
        \end{tabular}
    }
    \caption{Comparison with specialized or LLM-based NAS methods on NAS-Bench-201. The result of Random Search was reported by~\cite{lukasik2022learning}.}
    \label{tab:nas_bench_201}
\end{table}

\begin{table}[t]
    \centering
    \resizebox{0.95\linewidth}{!}{
        \begin{tabular}{l|c|c|c|c|c|c}
          \toprule
          \multirow{2}{*}{\textbf{Method}} & \multicolumn{2}{c|}{\textbf{CIFAR10}} & \multicolumn{2}{c|}{\textbf{CIFAR100}} & \multicolumn{2}{c}{ \textbf{ImageNet16-120}} \\

          & Pearson coe & P-value & Pearson coe & P-value & Pearson coe & P-value  \\
          \midrule
          NCode similarity & -0.0305 & 0.2953 & -0.0216 & 0.3152 & -0.0434 & 0.2829 \\
          \rowcolor[gray]{.9}Performance & -0.3591 & 0.0000 & -0.4022 & 0.0003 & -0.3082 & 0.0001 \\
          \bottomrule
        \end{tabular}
    }
    \caption{Correlation coefficients between attention score and architecture attributes.}
    \label{tab:attn_coe}
\end{table}

\end{document}